\documentclass[11pt,a4paper]{article}
\usepackage[hyperref]{acl2019}

\usepackage{times}
\usepackage{url}
\usepackage{latexsym}
\usepackage{xspace}
\usepackage{booktabs}
\usepackage{color}
\usepackage{graphicx}
\usepackage{tabulary}
\usepackage{enumitem}
\usepackage{url}
\usepackage{amsfonts}

\newenvironment{tightitemize}%
  {\begin{itemize}[topsep=0pt, partopsep=0pt] %
    \setlength{\itemsep}{0pt}%
    \setlength{\parskip}{0pt}%
    }%
  {\end{itemize}}
  \usepackage{booktabs}
\usepackage{subcaption}
\usepackage{lipsum}

  \usepackage{color}
  {\begin{enumerate}â \footnotesize%
    \setlength{\itemsep}{0pt}%
    \setlength{\parskip}{0pt}%
    }%
  {\end{enumerate}}


\usepackage{microtype}
\usepackage{multirow}
\usepackage{verbatim}
\usepackage{amsmath,amsthm,amssymb}
\usepackage{array}
\usepackage[scaled=0.86]{helvet}
\usepackage{ifthen}
\usepackage{courier}
\usepackage[boxed]{algorithm}
\usepackage{caption}
\usepackage{algpseudocode}
\aclfinalcopy

\belowdisplayskip \abovedisplayskip

\usepackage{amsmath}
\newcommand{\rel}{\textsc{rel}\xspace}

\newcommand{\none}{\textsc{none}~}

\newcommand{\loc}{\textsc{loc}}
\newcommand{\per}{\textsc{per}}
\newcommand{\org}{\textsc{org}}

\title{Entity-Relation Extraction as Multi-turn Question Answering}
\author{Xiaoya Li$^{*1}$, Fan Yin$^{*1}$, Zijun Sun$^1$, Xiayu Li$^1$\\ {\bf Arianna Yuan$^{1,3}$, Guoyin Wang$^2$, Duo Chai$^1$, Mingxin Zhou$^1$ and Jiwei Li$^1$ }\\
~~\\$^1$ Shannon.AI, $^2$ Amazon
~~\\$^3$ Computer Science Department, Stanford University
}

\date{}

\begin{document}
\maketitle
\begin{abstract}
  In this paper, we propose a new paradigm for the task of entity-relation extraction. 
  We 
 cast the task as a multi-turn question answering problem, i.e., the extraction of entities and relations is transformed to the task of identifying answer spans from the context. 
     This multi-turn QA formalization comes with several key advantages: firstly, the question query encodes important information for the entity/relation class we want to identify;
  secondly, QA provides a natural way of jointly modeling entity and relation; and thirdly, it allows us to exploit the well developed machine reading comprehension  (MRC) models. 
  
Experiments on the ACE and the CoNLL04 corpora demonstrate that the proposed paradigm significantly outperforms previous best models. 
We are able to obtain the state-of-the-art results on all of the ACE04, ACE05 and CoNLL04 datasets, increasing the SOTA results on the three datasets to 49.4 (+1.0), 60.2 (+0.6) and 68.9 (+2.1), respectively. 

Additionally, we construct and will release a newly developed dataset RESUME,  
which requires multi-step reasoning to construct entity dependencies, as opposed to the single-step dependency extraction in the triplet exaction in previous datasets. 
The proposed multi-turn QA model also achieves the best performance on the RESUME dataset. \footnote{* indicates equal contribution.}
\end{abstract}

\section{Introduction}
Identifying entities and their relations is the prerequisite of extracting structured knowledge from unstructured raw texts, which has recieved growing interest these years. 
Given a chunk of natural language text, the goal of entity-relation extraction is to transform it to a structural knowledge base. For example, given the following text:

{\it 
In  2002, Musk founded SpaceX, an aerospace manufacturer and space transport services Company, of which he is CEO and lead designer. He helped fund Tesla, Inc., an electric vehicle and solar panel manufacturer, in 2003, and became its CEO and product architect. In 2006, he inspired the creation of SolarCity, a solar energy services Company, and operates as its chairman.  In  2016, he co-founded Neuralink, a neurotechnology Company focused on developing brain–computer interfaces, and is its CEO. In  2016, Musk founded The Boring Company, an infrastructure and tunnel-construction Company.}

We need to extract four different types of entities, i.e., Person, Company, Time and Position, and three types of relations, \textsc{found}, \textsc{founding-Time} and \textsc{serving-role}.
The text is to be 
transformed into a structural dataset  shown in Table \ref{illustration}. 
\begin{table}
\small
\begin{tabular}{cccc}\hline
Person& Corp& Time & Position \\\hline
Musk& SpaceX & 2002 & CEO \\
\multirow{ 2}{*}{Musk}&\multirow{ 2}{*}{Tesla} &\multirow{ 2}{*}{ 2003} & CEO\&   \\
& & & product architect \\
Musk & SolarCity & 2006 & chairman  \\
Musk & Neuralink  & 2016 & CEO \\
Musk & The Boring Company & 2016 & - \\\hline
\end{tabular}
\caption{An illustration of an extracted structural table. }
\label{illustration}
\end{table}

Most existing models approach this task by extracting a list of triples from the text, i.e., \textsc{rel($e_1, e_2$)}, which denotes that 
 relation \rel holds
  between entity $e_1$ and entity $e_2$.  
  Previous models fall into two major categories: 
the pipelined approach, which  first uses tagging models to identify entities, and
then uses relation extraction models  to identify the relation between each entity pair; and
the joint approach, which
combines the entity model and the relation model throught different strategies, 
such as 
 constraints or
 parameters sharing. 

There are several key issues with current approaches, both in terms of the task formalization and the algorithm. 
At the formalization level, the \textsc{rel($e_1, e_2$)} triplet structure is not enough to fully express the data structure behind the text. 
Take the {\it Musk} case as an example, there is a hierarchical dependency between the tags: 
the extraction of Time depends on Position since a Person can hold multiple Positions in a Company during different Time periods. 
The extraction of Position also depends on Company since a Person can work for multiple companies. 
At the algorithm level, 
for most existing relation extraction models \cite{miwa2016end,wang2016relation,ye2016jointly}, 
the input to the  model is a raw sentence with two marked mentions, and the output is whether a relation holds between the two mentions. 
As pointed out in \newcite{wang2016relation,zeng2018extracting}, it is hard for neural models to capture all the lexical, semantic and syntactic cues in this formalization, especially when 
(1) entities are far away; 
(2) one entity is involved in multiple triplets;  or (3) relation spans have overlaps\footnote{e.g., in text {\it A B C D}, (A, C) is a pair and (B, D) is a pair.}.

In the paper, we propose a  new paradigm to handle the task of entity-relation extraction. We formalize the task as a multi-turn question answering task: 
each entity type and relation type is characterized by a question answering template, and entities and relations are extracted by answering template questions. 
Answers  are text spans, extracted using the now 
standard machine reading comprehension (MRC) framework: predicting answer spans given context \cite{seo2016bidirectional,wang2016machine,xiong2017dcn,wang2016multi}. 
To extract structural data like Table \ref{illustration}, the model need to answer the following questions sequentially:
\begin{tightitemize}
\item {\it Q: who is mentioned in the text? A: Musk};
\item {\it Q: which Company / companies did Musk work for?  A: SpaceX, Tesla, SolarCity, Neuralink and The Boring Company}; 
\item {\it Q: when did Musk join SpaceX?  A: 2002};
\item  {\it Q: what was Musk's Position in SpaceX?  A: CEO}.
\end{tightitemize}

Treating the entity-relation extraction task as a multi-turn QA task has the following key advantages: 
(1) the multi-turn QA setting provides an elegant way to capture the hierarchical dependency of tags. As the multi-turn QA proceeds, we progressively obtain the entities we need for the next turn. This is closely akin to the multi-turn slot filling dialogue system \cite{williams2005scaling,lemon2006isu};
(2) the question query encodes important prior information for the 
relation
 class we want to identify. For example, 
  information 
 in the query of the \textsc{PER} tagging class {\it who is mentioned in the text} 
helps the model to 
extract 
 relevant name entities.
On the contrary, in traditional non-QA entity-relation extraction models, a tagging classes or relation classes are merely indices (class1, class2, ...) and do not encode any information about the class.
This informativeness can potentially solve the issues that existing relation extraction models fail to solve, such as distantly-separated entity pairs,  relation span overlap, etc; 
  (3) the QA framework provides a  natural way to simultaneously extract entities and relations:  
most MRC models support outputting special  \textsc{None} tokens, indicating that there is no answer to the question. 
Throught this, the original two tasks, entity extraction and relation extraction 
 can be merged to a single QA task: 
a relation 
holds if the returned answer to the question corresponding to that relation 
 is not \textsc{None}, and this returned answer is the entity that we wish to extract.

In this paper, we show that the proposed  paradigm, which 
transforms
 the entity-relation extraction task to a multi-turn QA task,
    introduces significant performance boost over existing systems. It achieves state-of-the-art (SOTA) performance on the 
ACE and the CoNLL04 datasets.
The tasks on these datasets  are formalized as triplet extraction problems, in which two turns of QA suffice.
We thus build a more complicated and more difficult dataset called RESUME which requires to extract biographical information of individuals from raw texts. 
The construction of structural knowledge base from RESUME 
requires 
 four or five turns of QA. 
 We also show that this multi-turn QA setting could easilty integrate reinforcement learning (just as in multi-turn dialog systems) to gain additional performance boost. 
 
The rest of this paper is organized as follows: Section 2 details related work. We describe the dataset and setting in Section 3, the proposed model in Section 4, 
and experimental results in Section 5. We conclude this paper in Section 6. 
\section{Related Work}
\subsection{Extracting Entities and Relations}
Many  earlier entity-relation extraction systems are pipelined \cite{zelenko2003kernel,miwa2009rich,chan2011exploiting,lin2016neural}: an entity extraction model first identifies entities of interest and 
a relation extraction  model then constructs relations between the extracted entities. 
Although pipelined systems has the flexibility of integrating different data sources and learning algorithms, they suffer significantly from error propagation. 

To tackle this issue,
 joint learning models have been proposed. 
Earlier joint learning approaches
connect the two models through various dependencies, including 
constraints solved by integer linear programming \cite{yang2013joint,roth2007global}, card-pyramid parsing \cite{kate2010joint}, and global probabilistic graphical models \cite{yu2010jointly,singh2013joint}. 
In later studies, \newcite{li2014incremental} 
extract entity mentions and relations using structured perceptron with efficient beam-search, which is significantly more efficient and less Time-consuming than constraint-based approaches. 
\newcite{miwa2014modeling,gupta2016table,zhang2017end} 
proposed the  table-filling approach,
which provides an opportunity to incorporating 
more sophisticated features and algorithms into the model, such as search orders in decoding and global features.
Neural network models have been widely used in the literature as well. \newcite{miwa2016end} introduced an end-to-end approach that extract entities and their relations using neural network models with shared parameters, i.e., extracting entities using a neural tagging model and extracting relations using a neural multi-class classification model based on tree LSTMs \cite{tai2015improved}.  
\newcite{wang2016relation} extract relations using multi-level attention CNNs. 
\newcite{zeng2018extracting} proposed a new framework that uses sequence-to-sequence models to generate entity-relation triples, naturally combining entity detection and relation detection. 

Another way to bind the entity and the relation extraction models is to use reinforcement learning or Minimum Risk Training, in which the training signals are given based on the joint decision by the two models. \newcite{sun2018extracting}  
 optimized a global loss function to jointly train the two models under the framework work of Minimum Risk Training. \newcite{takanobu2018hierarchical} used hierarchical reinforcement learning to extract entities and relations in a hierarchical manner. 

\subsection{Machine Reading Comprehension}
Main-stream MRC models \cite{seo2016bidirectional,wang2016machine,xiong2017dcn,wang2016multi}
 extract text spans in passages given queries. Text span extraction can be simplified to two multi-class classification tasks, i.e., predicting the starting and the ending positions of the answer.  
Similar strategy can be extended to multi-passage MRC \cite{joshi2017triviaqa,dunn2017searchqa} 
 where the answer needs to be selected from multiple passages.
Multi-passage MRC tasks can be easily simplified to single-passage MRC tasks by concatenating passages 
\cite{shen2017reasonet,wang2017gated}. \newcite{wang2017evidence}
first rank the passages and then run single-passage MRC on the selected passage. 
\newcite{tan2017s} train the passage ranking model jointly with the reading comprehension model. 
Pretraining methods like BERT \cite{devlin2018bert} or Elmo \cite{peters2018deep} have proved to be extremely helpful in MRC tasks. 

There has been a tendency of casting 
non-QA NLP tasks as QA tasks \cite{mccann2018natural}. 
Our work is highly inspired by \newcite{levy2017zero}. \citet{levy2017zero} and \citet{mccann2018natural} focus on identifying the relation between two pre-defined entities and the authors formalize the task of relation extraction as a single-turn QA task. 
 In the current paper we study a more complicated scenario, where hierarchical tag dependency needs to be modeled and single-turn QA approach no longer suffices. 
  We show that our multi-turn QA method is able to solve this challenge and obtain new state-of-the-art results. 

\section{Datasets and Tasks}
\subsection{ACE04, ACE05 and CoNLL04}
We use ACE04, ACE05
and CoNLL04 \cite{roth2004linear},
the widely used entity-relation extraction benchmarks for evaluation. 
ACE04 defines 7  entity types, including  Person (\textsc{per}), Organization (\textsc{org}), Geographical Entities (\textsc{gpe}), Location (loc),
Facility (\textsc{fac}), Weapon (\textsc{wea}) and Vehicle (\textsc{veh}). 
For each pair of entities, it defines 7 relation categories, including 
Physical (\textsc{phys}), Person-Social (\textsc{per-soc}), Employment-Organization (\textsc{emp-org}), Agent-Artifact (\textsc{art}), PER/ORG Affiliation (\textsc{other-aff}), GPE-
Affiliation (\textsc{gpe-aff}) and Discourse (\textsc{disc}). 
ACE05 
 was built upon ACE04. It kept the  \textsc{per-soc}, \textsc{art} and \textsc{gpe-aff} categories from ACE04 but split \textsc{phys} into \textsc{phys}  and a new relation category 
\textsc{part-whole}. It also deleted \textsc{disc} and merged \textsc{emp-org} and \textsc{other-aff} into a new category \textsc{emp-org}. 
As for CoNLL04, it defines four entity
types (\loc, \org, \per and \textsc{others})  and five relation categories (\textsc{located\_in}, \textsc{work\_for}, \textsc{orgBased\_in}, \textsc{live\_in} ]and \textsc{kill}).

For ACE04 and ACE05, we followed the training/dev/test split in \newcite{li2014incremental} and \newcite{miwa2016end}\footnote{\url{https://github.com/tticoin/LSTM-ER/.}}. For the CoNLL04 dataset, we followed  \newcite{miwa2014modeling}.

\subsection{RESUME: A newly constructed dataset}
The ACE and the CoNLL-04 datasets are intended for triplet extraction, and two turns of QA is sufficient to extract the triplet (one turn for head-entities and another for joint extraction of tail-entities and relations).
These datasets do not involve hierarchical entity relations as in our previous {\it Musk} example, which are prevalent in real life applications.

Therefore, we construct a new dataset called RESUME.
We extract 841 paragraphs 
 from chapters describing management teams in IPO prospectuses.
 Each paragraph  
 describes some work history of an executive. We wish to extract the structural data from the resume. 

We identify four types of entities: Person (the name of the executive), Company (the company that the executive works/worked for), Position (the position that he/she holds/held) and Time (the time period that the executive occupies/occupied that position). 
It is worth noting that one person can work for different companies during different periods of time and that one person can hold different positions in different periods of time for the same company. 

We recruited crowdworkers to fill the slots in Table \ref{illustration}. 
We asked them to spend 5 
minutes on each passage and paid them \$$1$ per sentence.
Each passage is labeled by two different crowdworkers. 
If labels from the two annotators disagree, one or more annotators were asked to label the sentence and a majority vote was taken as the final decision.
Since the wording of the text is usually very explicit and formal, the inter-agreement between annotators is very high, achieving a value of 93.5\% for all slots. 
 Some statistics of the dataset are shown in Table \ref{resume}. We randomly split the dataset into training (80\%), validation(10\%) and test set (10\%). 
\begin{table}
\center
\small
\begin{tabular}{lcc} \hline
& Total \#  & Average \# per passage\\
Person&961 &1.09  \\
Company&1988  &2.13 \\
Position&2687 & 1.33 \\
Time&1275  & 1.01 \\\hline
\end{tabular}
\caption{Statistics for the RESUME dataset.}
\label{resume}
\end{table}

\section{Model}
\algrenewcommand{\algorithmicrequire}{\textbf{Input:}}
\algrenewcommand{\algorithmicensure}{\textbf{Output:}}
\newcommand{\To}{{\bf to }}
\newcommand{\IF}{{\bf if }}
\newcommand{\DO}{{\bf do }}
\newcommand{\ENDIF}{{\bf endif}}
\begin{algorithm}[t]
\small
\begin{algorithmic}[1]
\Require sentence $s$, 
 EntityQuesTemplates, ChainOfRelTemplates
\Ensure a list of list (table) M = []
\State
\State $M \gets \emptyset$
\State  HeadEntList$\gets \emptyset$
\For {entity\_question in EntityQuesTemplates}
  \State $e_1$ = Extract\_Answer(entity\_question, s)
  \State \IF $e_1\neq \textsc{None}$  \DO \\ 
  \hspace{0.7cm}  HeadEntList = HeadEntList +  $\{e_1\}$
   \State \ENDIF
 \EndFor
 \For {head\_entity in HeadEntList}
\State  ent\_list = [head\_entity]
 \For { [rel, rel\_temp] in ChainOfRelTemplates}
\For{(rel, rel\_temp) in List of [rel, rel\_temp]} 
\State q = GenQues(rel\_temp, rel, ent\_list)
     \State $e$ = Extract\_Answer(rel\_question, s)
       \State \IF $e\neq \textsc{None}$  
     \State\hspace{0.3cm}   ent\_list = ent\_list + e
     \State \ENDIF
     
\EndFor
 \EndFor
 \State \IF len(ent\_list)$= $len([rel, rel\_temp])
  \State\hspace{0.3cm} M = M + ent\_list 
 \State\ENDIF
 \EndFor
 \State \Return $M$
\end{algorithmic}
\caption{Transforming the entity-relation extraction task to a multi-turn QA task.}
\label{alg}
\end{algorithm}

\subsection{System Overview}

\begin{table*}[!ht]
\small
\center
\begin{tabular}{llll}\hline
Relation Type &head-e & tail-e&Natural Language Question \& Template Question  \\\hline
\textsc{gen-aff} & FAC & GPE& find  a  geo-political  entity that connects  to  XXX   \\
& & & XXX; has affiliation; geo-political entity \\\hline
\textsc{part-whole}& FAC& FAC& find  a  facility  that geographically  relates  to  XXX\\
& & & XXX; part whole; facility \\\hline 
\textsc{part-whole}& FAC& GPE&find  a  geo-political  entity  that geographically  relates  to  XXX  \\
& & & XXX; part whole; geo-political entity \\\hline 
\textsc{part-whole}& FAC & VEH & find  a  vehicle  that belongs  to  XXX   \\
& & & XXX; part whole; vehicle \\\hline 
\textsc{phys}&FAC & FAC &find  a  facility  near  XXX?    \\
& & & XXX; physical; facility \\\hline 
\textsc{art}&GPE & FAC& find  a  facility  which  is  made  by  XXX\\
& & & XXX; agent artifact; facility \\\hline 
\textsc{art}&GPE & VEH& find  a  vehicle  which  is owned  or  used  by  XXX\\
& & & XXX; agent artifact; vehicle \\\hline 
\textsc{art}&GPE & WEA& find  a  weapon  which  is owned  or  used  by  XXX\\
& & & XXX; agent artifact; weapon \\
\textsc{org-aff}&GPE & ORG&  find  an  organization  which  is  invested  by  XXX \\
& & & XXX; organization affiliation; organization \\\hline 
\textsc{part-whole}&GPE & GPE&   find  a  geo political  entity  which  is  controlled  by  XXX\\
& & & XXX; part whole; geo-political entity \\\hline 
\textsc{part-whole}&GPE & LOC&   find  a  location  geographically  related  to  XXX\\
& & & XXX; part whole; location \\\hline 
\hline 
\hline
\end{tabular}
\caption{Some of the question templates for different relation types in AEC. }
\label{reltemplate}
\end{table*}

\begin{table*}[!ht]
\center
\begin{tabular}{lll}\hline
Q1 Person:& who is mentioned in the text? &A: $e_1$ \\
Q2 Company:& which companies did $e_1$ work for? & A: $e_2$ \\
Q3 Position:& what was $e_1$'s position in $e_2$? & A: $e_3$ \\
Q4 Time:& During which period did $e_1$ work for $e_2$ as $e_3$ & A: $e_4$ \\\hline
\end{tabular}
\caption{Question templates for the RESUME dataset. }
\label{resumetem}
\end{table*}

The overview of the algorithm is shown in  Algorithm \ref{alg}.
The algorithm contains two stages: 

 (1) The head-entity extraction stage (line 4-9):
 each episode of multi-turn QA is triggered by an entity. 
 To extract this starting entity, 
  we transform 
    each entity type to a question using EntityQuesTemplates (line 4) and the entity $e$ is extracted by answering the question (line 5).
   If the system outputs the special \none  token, then it means $s$ does not contain any entity of that type.
   
    (2) The relation and the tail-entity extraction stage (line 10-24): ChainOfRelTemplates defines a chain of relations, the order of which we need to follow to run multi-turn QA.  
The reason is that the extraction of some entities depends on the extraction of others. For example, in the RESUME dataset, 
 the position held by an executive relies on the company he works for. Also the extraction of the Time entity relies on the extraction of both the Company and the Position.  
The extraction order is manually pre-defined. 
ChainOfRelTemplates also defines the template for each relation.
Each template contains some slots to be filled. 
 To generate a question (line 14), we insert previously extracted entity/entities to the slot/slots in a template. 
 The relation \rel and tail-entity $e$ will be jointly extracted by answering the generated question (line 15).
    A returned \none token indicates that there is no answer in the given sentence.

It is worth noting that entities extracted from the head-entity extraction stage may not all be head entities. 
In the subsequent relation and tail-entity extraction stage,   extracted entities from the first stage are initially assumed to be head entities, and are fed to the templates to generate questions. If an entity $e$ extracted from the first stage is indeed a head-entity of a relation, then the QA model will extract the tail-entity by answering the corresponding question. Otherwise, the answer will be \textsc{None} and thus ignored. 

For ACE04, ACE05 and CoNLL04 datasets, only two QA turns are needed. ChainOfRelTemplates thus only contain chains of 1.
For RESUME, we need to extract 4 entities, so ChainOfRelTemplates contain chains of 3. 

\subsection{Generating Questions using Templates}
Each entity type is associated with a type-specific question generated by the templates. 
There are two ways to generate questions based on templates: 
natural language questions or pseudo-questions. A pseudo-question is not necessarily grammatical. 
For example, the natural language question for the Facility type could be  {\it Which  facility  is  mentioned  in  the  text}, and the 
pseudo-question could just be {\it entity: facility}. 

At the relation and the tail-entity joint extraction stage, a question is generated 
by combing a relation-specific template with 
 the extracted head-entity.
 The question could be either a natural language question or a 
pseudo-question. 
Examples are shown in Table \ref{reltemplate} and Table \ref{resumetem}.

\subsection{Extracting Answer Spans via MRC}
Various MRC models have been proposed, such as BiDAF \cite{seo2016bidirectional} and QANet \cite{yu2018qanet}. 
In the standard MRC setting, given a question $Q=\{q_1, q_2, ..., q_{N_q}\}$ where $N_q$ denotes the number of words in $Q$, and 
context $C=\{c_1, c_2, ..., c_{N_c}\}$, where $N_c$ denotes the number of words in $C$, we need to predict the answer span. 
For the QA framework, we use BERT \cite{devlin2018bert} as a backbone.
BERT performs 
bidirectional language model pretraining on large-scale datasets using transformers \cite{vaswani2017attention} and achieves SOTA results on MRC datasets like SQUAD \cite{rajpurkar2016squad}. 
 To align with the BERT framework,  the question $Q$ and the context $C$ are combined  
 by concatenating the list [CLS, Q, SEP, C, SEP], where CLS and SEP are special tokens, $Q$ is the tokenized question and $C$ is the context. 
The representation of each context token is obtained using multi-layer transformers.

Traditional MRC models \cite{wang2016machine,xiong2017dcn} predict the starting and ending indices 
by applying
two softmax layers to the context tokens.
This
softmax-based span extraction strategy only fits for single-answer extraction tasks, but not for our task,  
since one sentence/passage in our setting might contain multiple answers. 
To tackle this issue, 
we formalize the  task as a query-based tagging problem \cite{lafferty2001conditional,huang2015bidirectional,ma2016end}. Specially, 
we predict a BMEO (beginning, inside, ending and outside) label for each token in the context given the query. 
The representation of each word is fed to a softmax layer to output a BMEO label. 
One can think that we are transforming two N-class classification tasks of predicting the starting and the ending indices 
(where $N$ denotes the length of sentence) to $N$ 5-class classification tasks\footnote{
For some of the relations that we are interested in, their corresponding questions have single answers.  We tried the strategy of predicting the starting and the ending index and found the results no different from the ones in the multi-answer QA-based tagging setting.  }.

\paragraph{Training and Test}
At the training time, we jointly train the objectives for the two stages:
\begin{equation}
\mathcal{L} = (1-\lambda) \mathcal{L}(\text{head-entity}) +\lambda \mathcal{L}(\text{tail-entity, rel}) 
\end{equation}
$\lambda\in [0,1]$ is the parameter controling the trade-off between the two objectives. Its value is tuned on the validation set. 
Both the two models are initialized using the standard BERT model and they share parameters during the training. 
At test time, head-entities and tail-entities are extracted separately based on the two objectives. 
\subsection{Reinforcement Learning}
Note that in our setting, 
the extracted answer from one turn 
not only affects its own accuracy, but also determines how
a question will be constructed for the downstream turns, which in turn affect later accuracies. We decide to use reinforcement learning to tackle it, which has been proved to be successful in  multi-turn dialogue generation \cite{mrkvsic2015multi,li2016dialogue,wen2016network}, a task that has the same challenge as ours.

\paragraph{Action and Policy} In a RL setting, we need to define action and policy.  
In the multi-turn QA setting, 
the action is selecting a text span in each turn.
The policy defines the probability of selecting a certain span given the question and the context. 
As the algorithm   relies on the BMEO tagging output, the probability of selecting a certain span $\{w_1, w_2, ..., w_n\}$ is the joint probability of $w_1$ being assigned to $B$ (beginning), $w_2, ..., w_{n-1}$ being assigned to $M$ (inside) and $w_n$ being assigned to $E$ (end), written as follows:
\begin{equation}
\begin{aligned}
&p(y(w_1, ..., w_n)=\text{answer}| \text{question}, s) \\ 
&=p(w_1 = \text{B})\times p(w_n = \text{E})\prod_{i\in[2,n-1]} p(w_i = \text{M})
\end{aligned}
\end{equation}
\paragraph{Reward}
For a given sentence $s$, we use the number of correctly retrieved triples as rewards. 
We use the REINFORCE algorithm \cite{williams1992simple}, a kind of policy gradient method, to find the optimal policy, which maximizes the expected reward $E_{\pi} [R(w)]$. The expectation is approximated by sampling from the policy $\pi$ and the gradient is computed using the likelihood ratio:
 \begin{equation}
\begin{aligned}
\nabla E(\theta)\approx [R(w)-b] \nabla\log \pi(y(w)|\text{question\ s}))
\end{aligned}
\label{hahaha}
\end{equation}
where $b$ denotes a baseline value.  
For each turn in the multi-turn QA setting, getting an answer correct leads to a reward of +1 . 
The final reward is the accumulative reward of all turns. 
The baseline value is set to the average of all previous rewards. 
We do not initialize policy networks from scratch, but use the pre-trained head-entity and tail-entity extraction model described in the previous section. 
We also use the experience replay strategy \cite{mnih2015human}: for each batch, half of the examples are simulated and the other half is randomly selected from previously generated examples. 

For the RESUME dataset, we use the strategy of curriculum learning \cite{bengio2009curriculum}, i.e., we gradually increase the number of turns from 2 to 4 at training. 

\begin{table*}
\center
\small
\begin{tabular}{|c|c|c|c|c|c|c|c|c|c|c|c|c|}\hline
& \multicolumn{3}{c|}{multi-turn QA} & \multicolumn{3}{c|}{multi-turn QA+RL} & \multicolumn{3}{c|}{tagging+dependency} & \multicolumn{3}{c|}{tagging+relation}\\\hline
& p & r & f& p & r & f & p & r & f & p & r & f\\\hline 
Person& {\bf98.1} & {\bf99.0} & {\bf98.6} & {\bf 98.1} & {\bf 99.0}  &{\bf98.6}         & 97.0&97.2& 97.1         & 97.0&97.2& 97.1\\
Company&82.3&87.6&84.9& {\bf 83.3} & {\bf 87.8} & {\bf 85.5}  & 81.4 &87.3 &   84.2            & 81.0&86.2 & 83.5\\
Position &97.1&98.5&97.8& {\bf 97.3} &{\bf 98.9}& {\bf 98.1} &    96.3 & 98.0 &     97.0       & 94.4 & 97.8&96.0\\
Time&96.6&98.8&97.7& {\bf 97.0} & {\bf 98.9}& {\bf 97.9}  &95.2 &96.3&     95.7               & 94.0& 95.9&94.9\\
all & 91.0&93.2&92.1&{\bf 91.6}&{\bf 93.5}&{\bf 92.5} &   90.0&91.7 &    90.8         &88.2&91.5 & 89.8\\\hline
\end{tabular}
\caption{Results for different models on the RESUME dataset.}
\label{resume-result}
\end{table*}

\begin{table*}
\begin{center}
\small
\begin{tabular}{lllccclll}\hline
Models& {\bf Entity P}&{\bf Entity R}&{\bf Entity F}&{\bf Relation P}&{\bf Relation R}&{\bf Relation F} \\\hline
\newcite{li2014incremental}&{\bf 83.5}&76.2&79.7&{\bf 60.8}&36.1&49.3 \\
\newcite{miwa2016end}&80.8&{\bf 82.9}&{\bf 81.8}&48.7&{\bf 48.1}&{\bf 48.4}  \\
\newcite{Katiyar2017}&81.2&78.1&79.6&46.4&45.3&45.7  \\
\newcite{D18-1307}&-&-&81.6&-&-&47.5  \\\hline
Multi-turn QA& {\bf 84.4} & 82.9 & {\bf 83.6}&50.1&{\bf 48.7}&{\bf 49.4} (+1.0) \\\hline
\end{tabular}
\end{center}
\caption{Results of different models on the ACE04 test set.  Results for pipelined methods are omitted since they consistently  underperform joint models (see \newcite{li2014incremental} for details).}
\label{ace04}
\end{table*}

\begin{table*}
\begin{center}
\small
\begin{tabular}{lllccclll}\hline
Models& {\bf Entity P}&{\bf Entity R}&{\bf Entity F}&{\bf Relation P}&{\bf Relation R}&{\bf Relation F} \\\hline
\newcite{li2014incremental}&{\bf 85.2} & 76.9 &80.8&{\bf 65.4}&39.8&49.5 \\
\newcite{miwa2016end}&82.9&{\bf 83.9}&83.4&57.2&54.0&55.6  \\
\newcite{Katiyar2017}&84.0&81.3&82.6&55.5&51.8&53.6  \\
\newcite{zhang2017end}&-&-&83.5&-&-&57.5  \\
\newcite{sun2018extracting} &83.9&83.2&{\bf 83.6}&64.9&{\bf 55.1}&{\bf 59.6}  \\\hline
Multi-turn QA& 84.7 & {\bf 84.9}& {\bf 84.8}&64.8&{\bf 56.2}&{\bf 60.2} (+0.6)\\\hline
\end{tabular}
\end{center}
\caption{Results of different models on the ACE05 test set. Results for pipelined methods are omitted since they consistently  underperform joint models (see \newcite{li2014incremental} for details).}
\label{ace05}
\end{table*}

\begin{table*}[!ht]
\begin{center}
\small
\begin{tabular}{lllccclll}\hline
Models &{\bf Entity P}&{\bf Entity R}&{\bf Entity F1}&{\bf Relation P} & {\bf Relation R} & {\bf Relation F} \\\hline
\newcite{miwa2014modeling} & -- & -- &80.7&--&--&61.0 \\
\newcite{zhang2017end}&--&--&85.6&--&--&67.8  \\
\newcite{D18-1307} &--&--&83.6&--&--&62.0  \\\hline
Multi-turn QA &{\bf 89.0}&{\bf 86.6}&{\bf 87.8}&{\bf 69.2}&{\bf 68.2}&{\bf 68.9} (+2.1) \\\hline
\end{tabular}
\end{center}
\caption{Comparison of the proposed method with the previous models on the CoNLL04 dataset. Precision and recall values of baseline models were not reported in the previous papers.}
\label{CoNLL04}
\end{table*}

\section{Experimental Results}
\subsection{Results on RESUME}
Answers are extracted according to the order of Person (first-turn), Company (second-turn), Position (third-turn) and Time  (forth-turn), and the extraction of each answer depends on those prior to them. 

For baselines, 
we first implement a joint model in which entity extraction and relation extraction are trained together (denoted by {\it tagging+relation}). As in \citet{zheng2017joint}, entities are extracted using BERT tagging models, and relations are extracted by applying a CNN to representations output by BERT transformers. 

Existing baselines which involve entity and relation identification stages (either pipelined or joint) are well suited for triplet extractions, but not really tailored to our setting because in the third and forth turn, we need more information to decide the relation than just the two entities. For instance, to extract Position, we need both Person and Company, and to extract Time, we need Person, Company and Position. 
This is akin to a  dependency parsing task, but at the tag-level rather than the word-level \cite{dozat2016deep,chen2014fast}. 
We thus proposed the following baseline, which modifies the previous entity+relation strategy to entity+dependency, denoted by {\it tagging+dependency}. 
We use the BERT tagging model to assign tagging labels to each word, and modify the current SOTA dependency parsing model Biaffine  \cite{dozat2016deep} to construct dependencies between tags. 
The Biaffine dependency model and the entity-extraction model are jointly trained. 

Results are presented in 
Table \ref{resume-result}. As can be seen, the tagging+dependency model outperforms the tagging+relation model. 
The proposed multi-turn QA model performs the best, with RL adding additional performance boost. 
Specially, for Person extraction, which only requires single-turn QA, the multi-turn QA+RL model performs the same as the multi-turn QA model. 
It is also the case in tagging+relation and tagging+dependency.

\subsection{Results on ACE04, ACE05 and CoNLL04} 
For ACE04, ACE05 and CoNLL04, only two turns of QA are required. 
For evaluation, 
we report micro-F1 scores, precision and recall on entities and relations (Tables \ref{ace04}, \ref{ace05}
and \ref{CoNLL04}) as in \newcite{li2014incremental,miwa2016end,Katiyar2017,zhang2017end}.
For ACE04,  the proposed multi-turn QA  model already outperforms previous SOTA by +1.8\% for entity extraction and +1.0\% for relation extraction, and the REINFORCE model adds an additional +0.2 gain for relation extraction. 
For ACE05, the proposed multi-turn QA  model outperforms previous SOTA by +1.2\% for entity extraction and +0.6\% for relation extraction, and the REINFORCE model adds an additional +0.1 for relation extraction. 
The proposed multi-turn QA model leads to a +2.2\% improvement on entity F1 and +1.1\% on relation F1.

\section{Ablation Studies}
\subsection{Effect of Question Generation Strategy}
In this subsection, we compare the effects of natural language questions and pseudo-questions. 
Results are shown in Table \ref{NLQ}. 
\begin{table}
\center
\small
\begin{tabular}{|lllllll|}\hline
\multicolumn{7}{|c|}{{\bf RESUME}}\\\hline
Model & \multicolumn{2}{|c|}{Overall P} & \multicolumn{2}{|c|}{Overall R} & \multicolumn{2}{|c|}{Overall F} \\\hline
{\bf Pseudo Q} &\multicolumn{2}{|c|}{90.2}&\multicolumn{2}{|c|}{92.3}&\multicolumn{2}{|c|}{91.2}\\
{\bf Natural Q} &\multicolumn{2}{|c|}{ 91.0}&\multicolumn{2}{|c|}{93.2}&\multicolumn{2}{|c|}{92.1} \\\hline\hline
\multicolumn{7}{|c|}{{\bf ACE04}}\\\hline
Model & EP & ER & EF & RP & RR & RF  \\\hline
{\bf Pseudo Q} &83.7 & 81.3 & 82.5 & 49.4 & 47.2 &48.3 \\
{\bf Natural Q} & {\bf 84.4} & {\bf 82.9} & {\bf 83.6}& {\bf 50.1} & {\bf 48.7} &{\bf 49.9}\\\hline\hline
\multicolumn{7}{|c|}{{\bf ACE05}}\\\hline
Model & EP &  ER &  EF & RP &  RR &  RF \\\hline
{\bf Pseudo Q} & 83.6 & 84.7 &84.2& 60.4 & 55.9 &58.1\\
{\bf Natural Q} & {\bf 84.7} & {\bf 84.9} & {\bf 84.8} & {\bf 64.8} & {\bf 56.2} &{\bf 60.2} \\\hline\hline
\multicolumn{7}{|c|}{{\bf CoNLL04}}\\\hline
Model & EP & ER & EF & RP & RR & RF \\\hline
{\bf Pseudo Q} & 87.4 & 86.4 & 86.9 & 68.2 & 67.4 &67.8 \\\hline
{\bf Natural Q}& {\bf 89.0} & {\bf 86.6} & {\bf 87.8} & {\bf 69.6} & {\bf 68.2} & {\bf 68.9} \\\hline
\end{tabular}
\caption{Comparing of the effect of natural language questions with   pseudo-questions.}
\label{NLQ}
\end{table}

We can see that natural language questions lead to a strict F1 improvement across all  datasets. 
This is because natural language questions provide more fine-grained semantic information and can help entity/relation extraction. By contrast, the pseudo-questions provide very coarse-grained,  ambiguous  and implicit hints of entity and relation types, which might even confuse the model.

\subsection{Effect of Joint Training}

In this paper, we decompose the entity-relation extraction task into two subtasks:  a multi-answer task for  head-entity extraction  
 and a single-answer task for joint relation and tail-entity extraction. 
 We jointly train two models with parameters shared. 
 The parameter $\lambda$ control the tradeoff between the two subtasks:
 \begin{equation}
\mathcal{L} =  (1-\lambda) \mathcal{L}(\text{head-entity}) +\lambda \mathcal{L}(\text{tail-entity}) 
\end{equation} 
Results regarding different values of $\lambda$ on the ACE05 dataset are 
given as follows: 
\begin{center}
\small
\begin{tabular}{ccccl}\hline
{\bf $\lambda$}&{\bf Entity F1}&{\bf Relation F1}\\\hline
$\lambda=0 $& 85.0&55.1 \\
$\lambda=0.1$& 84.8&55.4 \\
$\lambda=0.2$& {\bf 85.2}&56.2 \\
$\lambda=0.3$& 84.8&56.4 \\
$\lambda=0.4$& 84.6&57.9 \\
$\lambda=0.5$& 84.8&58.3 \\
$\lambda=0.6$& 84.6&58.9 \\
$\lambda=0.7$& 84.8&{\bf 60.2} \\
$\lambda=0.8$& 83.9&58.7 \\
$\lambda=0.9$& 82.7&58.3 \\
$\lambda=1.0$& 81.9&57.8 \\\hline
\end{tabular}
\end{center}

When $\lambda$ is set to 0, the system is essentially only trained on the head-entity prediction task. 
It is interesting to see that $\lambda=0$ does not lead to the best entity-extraction performance. 
This demonstrates that the second-stage relation extraction actually helps the first-stage entity extraction, which again confirms the necessity of considering these two subtasks together. 
For the relation extraction task, the best performance is obtained when $\lambda$ is set to 0.7.

\subsection{Case Study}
Table \ref{case} compares outputs from the proposed multi-turn QA model with the ones of the previous SOTA MRT model \cite{sun2018extracting}. 
In the first example, MRT is not able to identify the relation between {\it john scottsdale} and {\it iraq} because the two entities are too far away, but our proposed QA model is able to handle this issue.
In the second example, the sentence contains two pairs of the same relation. The MRT model has a hard time identifying handling this situation, not able to locate the {\it ship} entity and the associative relation, which the multi-turn QA model is able to  handle this case.

\begin{table}
\small
\begin{tabular}{ll}\hline
\textsc{example1} & [john scottsdale] \begin{tiny}PER: PHYS-1\end{tiny} is \\
&on the front lines in [iraq]\begin{tiny}GPE: PHYS-1\end{tiny} .  \\\hline
\textsc{MRT} & [john scottsdale] \begin{tiny}PER\end{tiny}\\
& is on the front lines in [iraq]\begin{tiny}GPE\end{tiny} .   \\\hline
\textsc{Multi-QA} & [john scottsdale] \begin{tiny}PER: PHYS-1\end{tiny} is \\
&on the front lines in [iraq]\begin{tiny}GPE: PHYS-1\end{tiny} .  \\\hline
\hline
\textsc{example2} & The [men]\begin{tiny} PER: ART-1\end{tiny} held on the\\
& sinking [vessel]\begin{tiny} VEH: ART-1\end{tiny} \\
& until the [passenger]\begin{tiny} PER: ART-2\end{tiny} \\
& [ship]\begin{tiny} VEH: ART-2\end{tiny} \\
& was able to reach them. \\\hline 
\textsc{MRT} & The [men]\begin{tiny} PER: ART-1\end{tiny} held on the \\
& sinking [vessel]\begin{tiny} VEH: ART-1\end{tiny} until \\
& the [passenger]\begin{tiny}PER\end{tiny}\\
& ship was able to reach them. \\\hline 
\textsc{Multi-QA} & The [men]\begin{tiny} PER: ART-1\end{tiny} held on the \\
& sinking [vessel]\begin{tiny} VEH: ART-1\end{tiny} \\
& until the [passenger]\begin{tiny} PER: ART-2\end{tiny} \\
& [ship]\begin{tiny} VEH: ART-2\end{tiny} was able to reach them. \\\hline 
\end{tabular}
\caption{Comparing the multi-turn QA model with MRT \cite{sun2018extracting}.}
\label{case}
\end{table}

\section{Conclusion}
In this paper, we propose a multi-turn question answering paradigm for the task of entity-relation extraction. We achieve new state-of-the-art results on 3 benchmark datasets. 
We also construct a new entity-relation extraction dataset that requires hierarchical relation reasoning and the proposed model achieves the best performance. 
\bibliography{acl2018}
\bibliographystyle{acl_natbib}

\end{document}